\documentclass[sigconf]{acmart}

\copyrightyear{2022}
\acmYear{2022}
\setcopyright{rightsretained}
\acmConference[KDD '22]{Proceedings of the 28th ACM SIGKDD Conference on Knowledge Discovery and Data Mining}{August 14--18, 2022}{Washington, DC, USA}
\acmBooktitle{Proceedings of the 28th ACM SIGKDD Conference on Knowledge Discovery and Data Mining (KDD '22), August 14--18, 2022, Washington, DC, USA}
\acmDOI{10.1145/3534678.3539173}
\acmISBN{978-1-4503-9385-0/22/08}

\usepackage{etoolbox}
\makeatletter
\patchcmd{\maketitle}{\@copyrightpermission}{
   \begin{minipage}{0.3\columnwidth}
     \href{https://creativecommons.org/licenses/by/4.0/}{\includegraphics[width=0.90\textwidth]{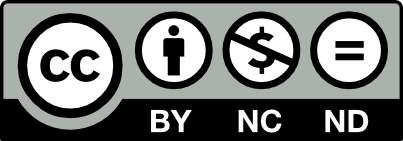}}
   \end{minipage}\hfill
   \begin{minipage}{0.7\columnwidth}
     \href{https://creativecommons.org/licenses/by-nc-nd/4.0/}{This work is licensed under a Creative Commons Attribution-NonCommercial-NoDerivs International 4.0 License.}
   \end{minipage}

   \vspace{5pt}
}{}{}

\makeatother

\usepackage{graphicx}
\usepackage{subfig}
\usepackage{url}
\usepackage{makecell}
\usepackage{balance} 
\usepackage{stfloats} 

\begin{document}

\title[Alexa Teacher Model]{Alexa Teacher Model: Pretraining and Distilling Multi-Billion-Parameter Encoders for Natural Language Understanding Systems}

\author{Jack FitzGerald}
\authornote{Corresponding Author - jgmf@amazon.com}
\affiliation{%
  \institution{Amazon, Denver, USA}
  \city{}
  \country{}
}

\author{Shankar Ananthakrishnan}
\affiliation{%
  \institution{Amazon, Cambridge, USA}
  \city{}
  \country{}
}

\author{Konstantine Arkoudas}
\affiliation{%
  \institution{Amazon, New York, USA} 
  \city{}
  \country{}
}

\author{Davide Bernardi}
\affiliation{%
  \institution{Amazon, Turin, Italy}
  \city{}
  \country{}
}

\author{Abhishek Bhagia}
\affiliation{%
  \institution{Amazon, Seattle, USA}
  \city{}
  \country{}
}

\author{Claudio Delli Bovi}
\affiliation{%
  \institution{Amazon, Turin, Italy}
  \city{}
  \country{}
}

\author{Jin Cao}
\affiliation{%
  \institution{Amazon, New York, USA}
  \city{}
  \country{}
}

\author{Rakesh Chada}
\affiliation{%
  \institution{Amazon, Seattle, USA}
  \city{}
  \country{}
}

\author{Amit Chauhan}
\affiliation{%
  \institution{Amazon, Seattle, USA}
  \city{}
  \country{}
}

\author{Luoxin Chen}
\affiliation{%
  \institution{Amazon, Cambridge, USA}
  \city{}
  \country{}
}

\author{Anurag Dwarakanath}
\affiliation{%
  \institution{Amazon, Bangalore, India}
  \city{}
  \country{}
}

\author{Satyam Dwivedi}
\affiliation{%
  \institution{Amazon, Bangalore, India}
  \city{}
  \country{}
}

\author{Turan Gojayev}
\affiliation{%
  \institution{Amazon, Aachen, Germany}
  \city{}
  \country{}
}

\author{Karthik Gopalakrishnan}
\affiliation{%
  \institution{Amazon, Santa Clara, USA}
  \city{}
  \country{}
}

\author{Thomas Gueudre}
\affiliation{%
  \institution{Amazon, Turin, Italy}
  \city{}
  \country{}
}

\author{Dilek Hakkani-Tur}
\affiliation{%
  \institution{Amazon, Sunnyvale, USA}
  \city{}
  \country{}
}

\author{Wael Hamza}
\affiliation{%
  \institution{Amazon, New York, USA}
  \city{}
  \country{}
}

\author{Jonathan Hueser}
\affiliation{%
  \institution{Amazon, Aachen, Germany}
  \city{}
  \country{}
}

\author{Kevin Martin Jose}
\affiliation{%
  \institution{Amazon, Aachen, Germany}
  \city{}
  \country{}
}

\author{Haidar Khan}
\affiliation{%
  \institution{Amazon, New York, USA}
  \city{}
  \country{}
}

\author{Beiye Liu}
\affiliation{%
  \institution{Amazon, Cambridge, USA}
  \city{}
  \country{}
}

\author{Jianhua Lu}
\affiliation{%
  \institution{Amazon, Cambridge, USA}
  \city{}
  \country{}
}

\author{Alessandro Manzotti}
\affiliation{%
  \institution{Amazon, Turin, Italy}
  \city{}
  \country{}
}

\author{Pradeep Natarajan}
\affiliation{%
  \institution{Amazon, Illinois, USA}
  \city{}
  \country{}
}

\author{Karolina Owczarzak}
\affiliation{%
  \institution{Amazon, Cambridge, USA}
  \city{}
  \country{}
}

\author{Gokmen Oz}
\affiliation{%
  \institution{Amazon, Cambridge, USA}
  \city{}
  \country{}
}

\author{Enrico Palumbo}
\affiliation{%
  \institution{Spotify, Turin, Italy}
  \city{}
  \country{}
}

\author{Charith Peris}
\affiliation{%
  \institution{Amazon, Cambridge, USA}
  \city{}
  \country{}
}

\author{Chandana Satya Prakash}
\affiliation{%
  \institution{Amazon, Cambridge, USA}
  \city{}
  \country{}
}

\author{Stephen Rawls}
\affiliation{%
  \institution{Amazon, New York, USA}
  \city{}
  \country{}
}

\author{Andy Rosenbaum}
\affiliation{%
  \institution{Amazon, Cambridge, USA}
  \city{}
  \country{}
}

\author{Anjali Shenoy}
\affiliation{%
  \institution{Amazon, Bangalore, India}
  \city{}
  \country{}
}

\author{Saleh Soltan}
\affiliation{%
  \institution{Amazon, New York, USA}
  \city{}
  \country{}
}

\author{Mukund Harakere Sridhar}
\affiliation{%
  \institution{Amazon, Cambridge, USA}
  \city{}
  \country{}
}

\author{Liz Tan}
\affiliation{%
  \institution{Amazon, Cambridge, USA}
  \city{}
  \country{}
}

\author{Fabian Triefenbach}
\affiliation{%
  \institution{Amazon, Aachen, Germany}
  \city{}
  \country{}
}

\author{Pan Wei}
\affiliation{%
  \institution{Amazon, Cambridge, USA}
  \city{}
  \country{}
}

\author{Haiyang Yu}
\affiliation{%
  \institution{Amazon, Cambridge, USA}
  \city{}
  \country{}
}

\author{Shuai Zheng}
\affiliation{%
  \institution{Amazon, Santa Clara, USA}
  \city{}
  \country{}
}

\author{Gokhan Tur}
\affiliation{%
  \institution{Amazon, Sunnyvale, USA}
  \city{}
  \country{}
}

\author{Prem Natarajan}
\affiliation{%
  \institution{Amazon, Los Angeles, USA}
  \city{}
  \country{}
}

\renewcommand{\shortauthors}{Jack FitzGerald et al.}

\begin{abstract}
We present results from a large-scale experiment on pretraining encoders with non-embedding parameter counts ranging from 700M to 9.3B, their subsequent distillation into smaller models ranging from 17M-170M parameters, and their application to the Natural Language Understanding (NLU) component of a virtual assistant system.
Though we train using 70\% spoken-form data, our teacher models perform comparably to XLM-R and mT5 when evaluated on the written-form Cross-lingual Natural Language Inference (XNLI) corpus.
We perform a second stage of pretraining on our teacher models using in-domain data from our system, improving error rates by 3.86\% relative for intent classification and 7.01\% relative for slot filling.
We find that even a 170M-parameter model distilled from our Stage 2 teacher model has 2.88\% better intent classification and 7.69\% better slot filling error rates when compared to the 2.3B-parameter teacher trained only on public data (Stage 1), emphasizing the importance of in-domain data for pretraining.
When evaluated offline using labeled NLU data, our 17M-parameter Stage 2 distilled model outperforms both XLM-R Base (85M params) and DistillBERT (42M params) by 4.23\% to 6.14\%, respectively.
Finally, we present results from a full virtual assistant experimentation platform, where we find that models trained using our pretraining and distillation pipeline outperform models distilled from 85M-parameter teachers by 3.74\%-4.91\% on an automatic measurement of full-system user dissatisfaction.
\end{abstract}

\ccsdesc[500]{Computing methodologies~Natural language processing}
\ccsdesc[500]{Human-centered computing~Personal digital assistants}
\ccsdesc[500]{Computing methodologies~Neural networks}

\keywords{natural language understanding, model pretraining, knowledge distillation, transformers, self-attention, distributed training, virtual assistant, voice a.i.}

\maketitle

\section{Introduction} \label{sect:intro}

A multi-step model training process is now dominant in most Natural Language Processing (NLP) applications, including Natural Language Understanding (NLU) \cite{devlin2019bert}.
In the first step of training, usually called pretraining, models are trained on large, self-supervised datasets, and they are tasked to fill in masked words, de-shuffle words or sentences, or predict the next word of a sequence.
In the final step, called fine-tuning, the model is adapted to a specific task using a comparatively small labeled dataset.
Additional training steps optionally can occur between the first pretraining step and the final fine-tuning step.

In parallel to this paradigm shift, researchers have discovered a clear correlation between task performance and model size, motivating work on dense models with tens or hundreds of billions of parameters and sparse models with up to trillions of parameters (see Section \ref{sect:related}).
To be useful for latency-sensitive online applications, large models must be distilled into smaller versions or otherwise compressed.
Recent knowledge distillation techniques have resulted in up to 99\% \cite{wang-etal-2021-minilmv2} to 96.8\% \cite{jiao-etal-2020-tinybert} of task performance preservation even after model size reductions of 50\% to 86\%, respectively.
Moreover, models distilled from larger models typically outperform models trained from scratch at the target size \cite{soltan-etal-2021-limitations}.

In this work we consider language model pretraining and distillation for improving the NLU performance of a large-scale virtual assistant. Our core tasks are intent classification and slot filling. Given the utterance ``\textit{can you call mom},'' the NLU model should understand that the user's intent is to make a call, and it should also fill the contact name slot with the ``mom'' token.

We refer to our models and pipeline as Alexa Teacher Model(s) (AlexaTM) throughout this paper.
Our problem space is unique as compared to many research tasks because (1) we possess relatively large labeled datasets, which reduces the effectiveness of pretraining, (2) our models must adhere to strict latency and memory constraints, (3) incoming data is of ``spoken form'' which differs from the ``written form'' text used to pretrain most public models, and (4) our system supports more than one language.

Our contributions include:
\begin{itemize}
\item{The first example (to our knowledge) of billion-parameter encoder pretraining using spoken-form data, as well as comparisons to models trained with written-form data,}
\item{Results from performing Stage 2 pretraining of the teacher models using in-domain data from a large, real-world system,}
\item{Setup and results for knowledge distillation to a student 0.2\% as large as its teacher (9.3B to 17M), contrasted, for example, with TinyBERT$_4$, which is 6\% the size of its teacher (85M to 5M),}
\item{Standalone results of our teacher and distilled models on both public datasets and datasets from a major NLU system, and}
\item{Full virtual assistant system results comparing our models to baseline models trained by smaller teachers.}
\end{itemize}

\section{Setup} \label{sect:setup}

\subsection{Pretraining Datasets} \label{sect:datasets}

Pretraining requires large datasets composed of diverse data spanning many domains, topics, tones, levels of formality, desired languages, and more.
We considered three primary pretraining data sources, being the multilingual Colossal Clean Common Crawl (mC4) dataset, which was used to train T5 \cite{raffel2020exploring} and mT5 \cite{xue2021mt5}, the CC-100 dataset, which was used to train XLM-R \cite{conneau-etal-2020-unsupervised, wenzek-etal-2020-ccnet}, and Wikipedia data, which was used to train BERT and mBERT in addition to the BooksCorpus \cite{devlin2019bert}.
mC4 and CC-100 are derived from Common Crawl data.

We included 12 languages for pretraining: Arabic, English, French, German, Hindi, Italian, Japanese, Marathi, Portuguese, Spanish, Tamil, and Telugu. Following \citep{NEURIPS2019_c04c19c2} we sampled sentences from the training corpus according to a multinomial distribution $\{q_i\}_{i=1\dots{}N}$, where:

\begin{equation}
q_i = \frac{p_i^\alpha}{\sum_{j=1}^{N}{p_j^\alpha}} \quad \textrm{with} \quad p_i = \frac{n_i}{\sum_{k=1}^N{n_k}} \textrm{,}
\end{equation}
$n_i$ is the number of examples in a given language's dataset, and $N$ is the number of languages considered.

We used $\alpha=0.5$ to mix data for both tokenizer training  and language model pretraining. The effect is to up-sample low-resource languages.
This up-sampling was performed offline prior to training.
Besides language-based sampling, we also packed sentences into sequences of approximately 700 words.
We performed tokenization on the fly during training, and 700 words per example allowed us to keep over 90\% of sequences under 1,024 tokens post-tokenization.

In addition to public datasets, we considered a proprietary Stage 2 pretraining dataset composed of unlabeled and anonymized utterance text from our system.
As preprocessing, we first reduced the duplication of examples in the dataset by repeating a given utterance only the square root of its actual count.
For instance, an example that appeared 100 times in the original dataset was reduced to appear only 10 times.
Second, we used the same language-sampling technique as described above for the Stage 1 pretraining dataset.
Third, we removed examples with a length of fewer than 5 tokens.
Finally, in order to reduce catastrophic forgetting, we then mixed the data with the public dataset used for Stage 1 pretraining following a 1:2 ratio of Stage 1 data to in-house data.
Our final Stage 2 pretraining dataset had approximately 50M examples.
Figure \ref{fig:distillation} shows how our datasets were used in our training pipeline.

\subsection{Spoken Form Text} \label{sect:spokenformpartone}

In Spoken Language Understanding (SLU) systems \citep{Young2002TalkingTM, Wang2005SpokenLU, Tur2011SpokenLU}, which are composed of both Automatic Speech Recognition (ASR) and NLU components, it is common to transform text from its original ``written form'' into a canonical ``spoken form'' to facilitate ASR.
This may include lower-casing, verbalizing of numbers, etc.
For example, a text like \emph{``Can you set an alarm for 7:30AM?''} might be converted to \emph{``can you set an alarm for seven thirty a. m.''}.

We observed that differences in the tokenization format can impact downstream performance. 
For instance, XLM-R Base, when trained on the written form of XNLI, has an English accuracy of 85.2, while on the spoken form of the English test set it drops to 83.4. 
We find such drops due to mismatched formatting to be more significant in smaller sized models like the ones used in a production system.
To mitigate this and better align with our use cases, we train on a mixed tokenization regime. 
To support both formats, yet bias towards the spoken-form setting, we transformed our pretraining data into spoken form using in-house formatters, and we mixed the spoken-form version (70\%) with the original written-form version (30\%).

\subsection{Tokenizer} \label{sect:tokenizer}

We trained a SentencePiece \cite{kudo-richardson-2018-sentencepiece} tokenizer using the unigram setting.
As shown in \cite{conneau-etal-2020-unsupervised}, tokenizer vocabulary size can have a large impact on model performance.
Larger vocabulary sizes generally lead to better task performance at the cost of training convergence speed, inference memory, and latency for masked language modeling.

It is prohibitively expensive to train a full teacher model with numerous tokenizer settings, so we developed two intrinsic tokenizer metrics: (1) split-ratio, and (2) unk-token portion.
Split-ratio uses the intuition that more subword splits will result in degraded accuracy.
Unk-token portion is defined as the percentage of output tokens that have the unknown token \texttt{<unk>}, which can seriously harm performance.

We increased the vocabulary size until we had a split-ratio and unk-token portion similar to our baseline production models (Section \ref{sect:onlineresults}).
To improve coverage for Japanese characters, we explicitly added the full set of the 2,136 of JōYō most common kanji \cite{enwiki:1039289460}, as well as all hiragana and katakana symbols.

We arrived at a vocabulary size of 150k subword tokens, and we used the same 70/30 mix of spoken-form and written-form data that we used for the pretraining corpus.

\subsection{Pretraining} \label{sect:pretrainingsetup}

We performed pretraining following the general examples of BERT \cite{devlin2019bert}, RoBERTa \cite{liu2019roberta}, XLM-R \cite{conneau-etal-2020-unsupervised}, and others.
Our teacher models are based on RoBERTa, but we modified them to use a pre-layernorm architecture, meaning that the layer normalization occurs immediately prior to the self attention block and the feedforward block in each transformer layer \cite{xiong2020layer}.

Training was conducted using the masked language modeling objective, in which 15\% of tokens are masked, of which 10\% are kept unchanged and 10\% are replaced with a random token.

We trained teacher models with up to 9.3B non-embedding parameters, and we used Deepspeed to increase our training throughput \cite{rajbhandari2020zero}. 
Deepspeed Stage 1 partitions optimizer states across GPUs, and Deepseed Stage 2 further partitions gradients across GPUs.
This partitioning can be achieved without any increase in network-based bottlenecks.
With mixed precision training, were able to achieve up to 107 TFLOP/sec per GPU for a 9.3B-parameter encoder using AWS p4d.24xlarge instances, which are composed of Nvidia a100 GPUs, using Elastic Fabric Adapters to ensure good network throughput.

We used Deepspeed's version of mixed precision training for our pretraining runs, and we encountered FP16 overflow during certain operations in the model. To mitigate these issues, we (1) used the \texttt{baddbmm} operation instead of the \texttt{matmul} operation for our query-key multiplication and (2) converted to FP32 prior to calculating the variance as part of the layer normalizations. These changes reduced throughput by up to 20\%, but they eliminated our model stability issues. Another way to mitigate the stability issues is to use BFLOAT16 \cite{kalamkar2019study}, which was not available in Deepspeed at the time of our experiments.

\subsection{Stage 2 Pretraining} \label{sect:phase2setup}

Stage 2 pretraining was explored with the Muppet system \cite{aghajanyan2021muppet}, which the authors of the associated paper refer to as pre-finetuning. Though their pre-finetuning is multitask, for our Stage 2 pretraining, we simply continued the pretraining objective using our Stage 2 dataset described in Section \ref{sect:datasets}. The goal was to improve our model's specialization and ability to handle virtual assistant utterances, which are typically short and often ungrammatical, while not catastrophically forgetting general language knowledge learned during Stage 1 \cite{cao2020style,gururangan2020dont}. Our hyperparameter choices are explained further in Section \ref{sect:phase2results}.

To examine the effectiveness of domain adaptation with Stage 2 pretraining, we evaluated our models' performance on intent classification and slot filling, using data from three diverse domains. Each domain's dataset contained a mix of 7 languages---English, German, Spanish, French, Italian, Portuguese and Japanese---and each dataset contained 80k to 90k training utterances. Statistics are shown in Table \ref{tab:stage2evaldata}.

\begin{table}[h]
\centering
\resizebox{\columnwidth}{!}{
\begin{tabular}{cccc}
\hline
 & \textbf{Domain 1} & \textbf{Domain 2} & \textbf{Domain 3} \\
 \hline
Training data size & 90k & 86k & 80k \\
Validation data size & 10k & 10k & 10k \\
Test data size & 20k & 20k & 20k \\
\# of intents & 16 & 8 & 12 \\
\# of slots & 98 & 25 & 56 \\
\hline
\end{tabular}
}
\caption{Description of the manually transcribed and labeled datasets used for offline NLU evaluation of Stage 2 versus Stage 1 performance.}
\label{tab:stage2evaldata}
\end{table}

We adopted two modes for Stage 2 evaluation. 
First, we followed a standard fine-tuning scheme, allowing all parameters, including parameters from the pretrained encoder as well as from classification head, to adapt to the task. 
Another mode we adopted is to freeze all the parameters from the pretrained network and only allow parameters in the classification head to learn.
We consider the latter mode as a more difficult task, given that the entire pretrained encoder is frozen and is essentially used as a feature extractor.
Thus, this latter mode may be a stronger indicator for the effectiveness of the pretrained encoder at creating generic representations useful for downstream tasks.

\subsection{Distillation} \label{sect:distillationsetup}

\begin{figure*}
	\centering
	\includegraphics[width=0.9\linewidth]{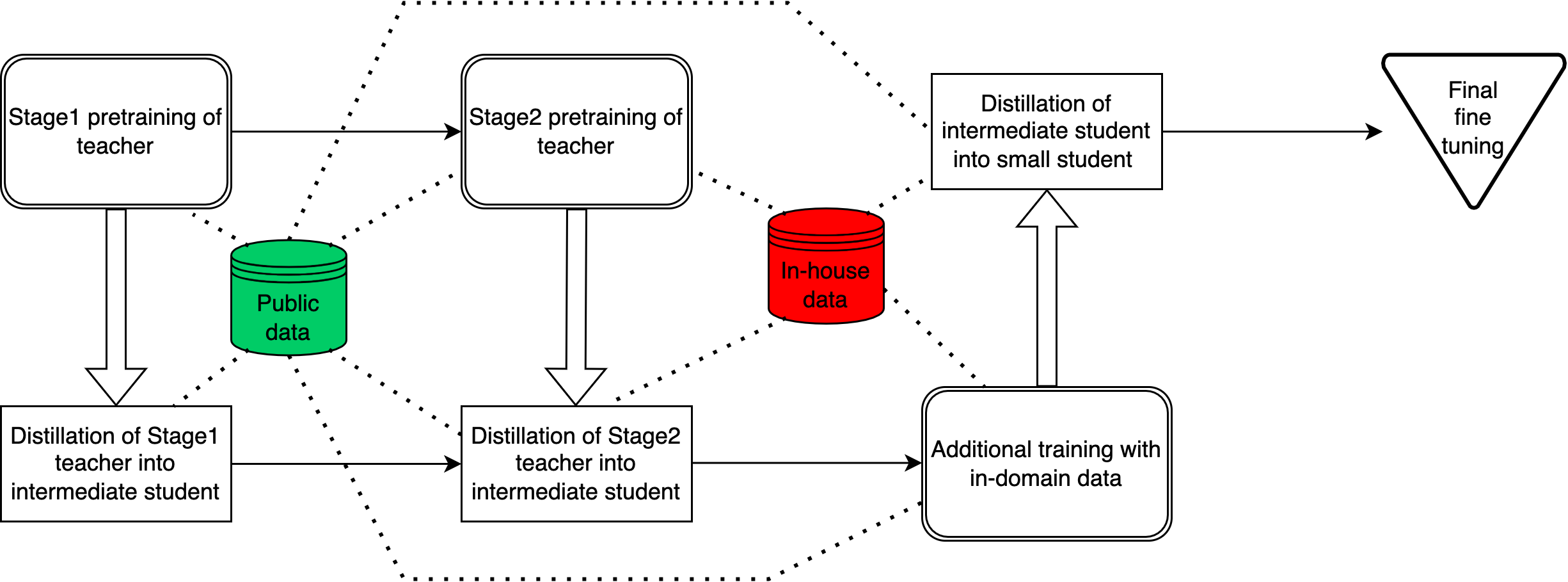}
	\caption{Our model training pipeline. A large teacher is first pretrained using public data as Stage 1. Pretraining continues with Stage 2 in-house data to create a new teacher. We then distill an intermediate student, starting with the Stage 1 teacher and then using the Stage 2 teacher. The intermediate student/teacher is then further trained on in-house unlabeled data before being distilled into the final student. The final student in then fine-tuned on labeled data.}
    \label{fig:distillation}
\end{figure*}

\begin{figure*}
    \centering
    \subfloat[\centering Perplexity]{{\includegraphics[width=0.35\linewidth]{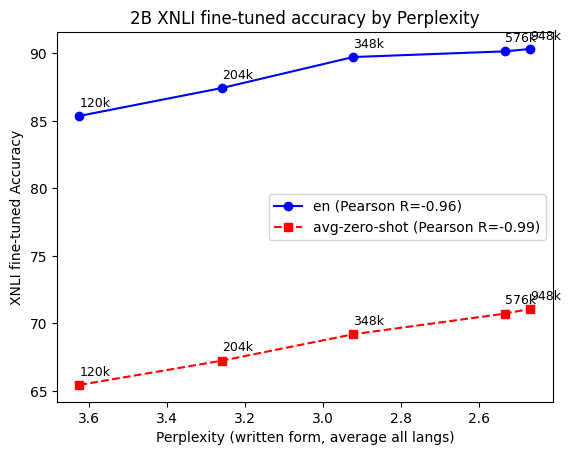} }}
    \qquad
    \subfloat[\centering Mask-filling Accuracy]{{\includegraphics[width=0.35\linewidth]{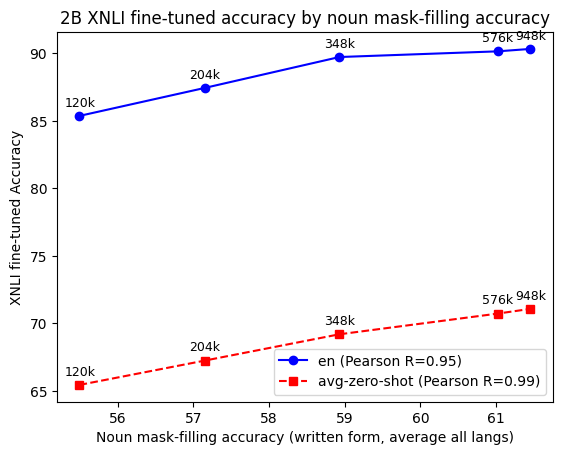} }}
    \caption{Correlation to XNLI accuracy from (a) perplexity and (b) mask-filling accuracy across model updates using our 2.3B-parameter model. The greater the correlation, the better the metric is for no-fine-tune validation.}
    \label{fig:xnlicorr}
\end{figure*}

Low-latency applications require models of relatively small sizes. 
However, distilling from large pretrained models into much smaller models directly can hinder the student from fully taking advantage of the teacher's knowledge \cite{cho2019efficacy}. 
Therefore, we distill the pretrained teacher models in two phases with a teacher assistant setup \cite{Mirzadeh_Farajtabar_Li_Levine_Matsukawa_Ghasemzadeh_2020, wang2020minilm}. 
The distillation workflow is depicted in Figure \ref{fig:distillation}. 
First, we distill an intermediate sized model from the large teacher model. We then use this distilled model as a teacher for the final student.

When distilling the intermediate model, we followed a similar approach to the pretraining of the teacher. 
A randomly initialized student model was distilled from the Stage 1 teacher model. 
Once training converged, we switched the teacher model to the Stage 2 teacher and resumed the distillation process.
For both of these stages, the distillation data is the same that was used for teacher pretraining for its respective stage.
As for our distillation techniques, we explored different components described in \cite{jiao-etal-2020-tinybert}.
Our final run for the intermediate student/teacher used the sum of categorical cross-entropy (MLM loss) and soft cross-entropy weighted equally, because we did not observe any gain from utilizing the attention and hidden layer outputs of the teacher.

For our final student, we first pretrained the intermediate model further without teacher involvement on Stage 2 data only. 
Next, we distilled it into the final, small student.
The distillation techniques in this phase were similar to the first distillation phase, with an additional usage of hidden-layer output matching as in \cite{jiao-etal-2020-tinybert}.

\begin{table*}[h]
\centering
\begin{tabular}{lccccccc}
\hline
\textbf{Model} & \textbf{en} & \textbf{ar} & \textbf{de} & \textbf{es} & \textbf{fr} & \textbf{hi} & \textbf{avg 0-shot}\\
\hline
XLM-R Base (0.27B) & 85.8 & 73.8 & 78.7 & 80.7 & 79.7 & 72.4 & 77.1 \\
XLM-R Large (0.6B) & 89.1 & 79.8 & 83.9 & 85.1 & 84.1 & 76.9 & 82 \\
XLM-R XL (3.5B) & 90.7 & 81.6 & 84.6 & 86.5 & 85.5 & 78.5 & 83.3 \\
XLM-R XXL (10.7B) & 91.6 & 82.5 & 87 & 87.3 & 86.2 & 79.8 & 84.6 \\
mT5 Large (1.2B) & 89.4 & 79.8 & 83.4 & 84.2 & 84.1 & 77.6 & 81.8 \\
mT5 XL (3.7B) & 90.6 & 82.2 & 85.8 & 81.3 & 85.3 & 80.4 & 83 \\
mT5 XXL (13B) & 92.3 & 84.4 & 87.3 & 88.3 & 87.3 & 82.5 & 86 \\
\hline
AlexaTM 9.3B Stage 1 (9.9B) & 91.9 & 82.2 & 86.9 & 87.4 & 86.8 & 80.2 & 84.7 \\
AlexaTM 2.3B Stage 1 (2.68B) & 90.3 & 80 & 84.7 & 85.9 & 85.3 & 77.3 & 82.6 \\
\hline
AlexaTM 170M from 2.3B Stage 1 (0.33B) & 87.3 & 77.6 &	81 & 82.5 &	81.7 &	74.6  & 79.5\\
\hline
\end{tabular}
\caption{Results on XNLI for the Stage 1 pretrained 2.3B- and 9.3B-parameter models, as well as the 170M-parameter model distilled from the Stage 1 2.3B-parameter model. The number of parameters including the embeddings is given in parentheses.}
\label{tab:phase1xnli}
\end{table*}

\subsection{No-Fine-Tune Validation} \label{sect:nftsetup}

In order to monitor the progress of training, one standard approach is to measure perplexity on a held-out validation dataset.
One issue with perplexity measurement is that it differs depending on the tokenizer choice.
Thus, we developed a separate task called ``mask-filling accuracy'' to compare models.

We selected texts from a variety of public tasks including XNLI \cite{conneau2018xnli}, PAWS-X \cite{yang2019pawsx}, and Multi-lingual Amazon Reviews \cite{keung2020multilingual}.
We then removed these examples from our training data.
For each example, we use the Stanza tagger \cite{qi2020stanza} to identify a noun word, then mask all subword tokens for that word.
The model must correctly predict all subword tokens in the noun to count as correct.

We show (Figure \ref{fig:xnlicorr}) that both perplexity and mask-filling accuracy correlate strongly with XNLI performance across model update steps.

\section{Results and Analyses} \label{sect:results}

\subsection{Stage 1 Pretraining} \label{sect:phase1results}

To measure the effectiveness of our pretraining and distillation setup using Stage 1 data (public data), we used XNLI to benchmark our 2.3B-parameter teacher model, a similarly trained 9.3B-parameter teacher model, and a 170M-parameter model distilled from the 2.3B-parameter model.
Following standard practice, we trained and validated on English data only, and we tested on all languages separately.
Non-English test results were averaged to determine the average zero-shot accuracy.
See Table \ref{tab:phase1xnli}.

We found that the 2.3B-parameter and 9.3B-parameter models are competitive with comparably-sized public models, even though our model training set was 70\% spoken-form data, whereas the public models and XNLI use written-form data. English XNLI accuracy drops by 3 points after distillation from 2.3B non-embedding parameters to 170M parameters, as well as by 3.1 points for average zero-shot accuracy.

Next, we examined the perplexity and noun mask-filling accuracy for spoken-form data derived from datasets spanning all of our languages, including XNLI, PAWS-X, and Amazon Reviews, as described in Section \ref{sect:nftsetup}. See Table \ref{tab:nftresults}. We expected to see perplexity decrease and noun mask filling accuracy increase with increasing model sizes, which we do observe.

\begin{table}[h]
\centering
\begin{tabular}{lcc}
\hline
 \textbf{Model}      & \textbf{Perplexity}   &   \textbf{Noun Mask Fill Acc} \\
\hline
XLM-R Large & 39.04 & 49.69 \\
AlexaTM 2.3B Stage 1 & 11.56 & 62.74 \\
AlexaTM 9.3B Stage 1 & 8.80 & 65.09 \\
\hline
\end{tabular}
\caption{No-fine-tune perplexity and noun mask-filling accuracy on spoken-form data only, macro-averaged across all languages. See Sections \ref{sect:nftsetup} and \ref{sect:pretrainingsetup}. Note that XLM-R was trained on written-form data.}
\label{tab:nftresults}
\end{table}

\subsection{Stage 2 Pretraining} \label{sect:phase2results}

\begin{table*}[h]
\centering
\subfloat[][]{
\resizebox{0.45\linewidth}{!}{
\begin{tabular}{lcccc}
\hline
 \multicolumn{5}{c}{Full Fine Tuning} \\
\hline
 \multicolumn{5}{c}{Relative Intent Class Error Reduction Versus 2.3B Stage 1} \\
 \hline
 & Domain 1 & Domain 2 & Domain 3 & Avg \\
 2.3B Stage 2 & -3.41\% & -2.38\% & -5.79\% & -3.86\% \\
170M from 2.3B & -3.16\% & -4.13\% & -1.36\% & -2.88\% \\
17M from 170M & 11.49\% & 10.63\% & 10.73\% & 10.95\% \\
\hline
\multicolumn{5}{c}{Relative Slot Filling Error Reduction Versus 2.3B Stage 1}\\
\hline
 & Domain 1 & Domain 2 & Domain 3 & Avg \\
 2.3B Stage 2 & -5.40\% & -9.95\% & -5.68\% & -7.01\% \\
170M from 2.3B & -2.52\% & -12.03\% & -8.53\% & -7.69\% \\
17M from 170M & 27.07\% & 2.11\% & 5.36\% & 11.51\% \\
\hline
\end{tabular}
}
}
\hfill
\subfloat[][]{
\resizebox{0.45\linewidth}{!}{
\begin{tabular}{lcccc}
\hline
 \multicolumn{5}{c}{Frozen Encoder} \\
 \hline
 \multicolumn{5}{c}{Relative Intent Class Error Improvement Versus 2.3B Stage 1} \\
\hline
 & Domain 1 & Domain 2 & Domain 3 & Avg \\
 2.3B Stage 2 & -12.60\% & -4.59\% & -2.23\% & -6.47\%  \\
 170M from 2.3B & -16.07\% & -17.23\% & -13.95\% & -15.75\%  \\
17M from 170M & 13.99\% & 7.42\% & 10.83\% & 10.74\% \\
\hline
\multicolumn{5}{c}{Relative Slot Filling Error Improvement Versus 2.3B Stage 1} \\
\hline
 & Domain 1 & Domain 2 & Domain 3 & Avg \\
2.3B Stage 2 & -5.51\% & -18.71\% & -6.72\% & -10.31\% \\
170M from 2.3B & -6.15\% & -12.03\% & -3.70\% & -7.29\% \\
17M from 170M & 15.41\% & -6.30\% & 3.20\% & 4.11\% \\
\hline
\end{tabular}
}
}
\caption{(a) Full fine-tuning and (b) frozen-encoder results for the 2.3B-parameter Stage 2 model, the distilled 170M-parameter Stage 2 model, and the 17M-parameter Stage 2 model, evaluated using a natural language understanding dataset (intent classification and slot filling) from our real-world system (see Table \ref{tab:stage2evaldata}). A negative value indicates a reduced error rate versus the baseline 2.3B-parameter Stage 1 model.}
\label{tab:stage2offline}
\end{table*}

\begin{table*}[h]
\centering
\begin{tabular}{lcccccccc}
\hline
 & Loc 1 & Loc 2 & Loc 3 & Loc 4 & Loc 5 & Loc 6 & Loc 7 & Avg \\
 \hline
Distill-mBERT & 5.50\% & 2.07\% & 1.61\% & -2.30\% & 1.41\% & 3.74\% & 12.34\% & 3.48\% \\
AlexaTM 170M Stage 2 & -2.20\% & -8.53\% & -7.61\% & -5.84\% & -7.64\% & -2.80\% & 0.90\% & -4.82\% \\
AlexaTM 17M Stage 2 & 0.50\% & -6.12\% & -6.19\% & -8.02\% & -5.64\% & -1.51\% & -2.63\% & -4.23\% \\
\hline
\end{tabular}
\caption{Exact match results for our AlexaTM distilled models and DistillBERT versus XLM-R. A given example is a successful exact match if the intent and all slots are correct. All models are trained on the same training sets as used for Section \ref{sect:onlineresults}. Results are given across 7 locales (language and region). A negative value indicates an improvement in exact match error rate.}
\label{tab:offlinenlu}
\end{table*}

To examine the effectiveness of Stage 2 pretraining, we used the 2.3B-parameter Stage 1 model as a baseline and compared it to various sizes of the Stage 2 models, including the 2.3B-parameter Stage 2 model, the 170M-parameter Stage 2 model distilled from the 2.3B-parameter Stage 2 model, and the 17M-parameter Stage 2 model distilled from the 170M-parameter Stage 2 model.
For classification heads, we implemented two feed-forward layers of hidden size 256, followed by one softmax layer for intent classification and one for slot filling.

Throughout our experiments, we trained all models (baselines and different size of Stage 2 models) with mini-batch sizes ranging from 16 to 64 using 8 Nvidia Tesla V100 GPUs. We used the Adam optimizer, a maximum learning rate of 2e-5 for the fine-tune mode, and a maximum learning rate of 1e-3 for the frozen mode . We report mean statistics across 3 random seed runs.

We show the domain adaptation results from full fine-tuning and the frozen encoder mode in Tables \ref{tab:stage2offline}(a) and \ref{tab:stage2offline}(b).
In fine-tune mode and compared with the 2.3B-parameter Stage 1 model, the 2.3B-parameter Stage 2 model reduces intent classification error rate by 3.86\% on average, as well as slot filling errors by 7.01\% on average. The 170M-parameter Stage 2 model performs surprisingly similarly to its 2.3B-parameter Stage 2 teacher, suggesting that the 2.3B-parameter model may be overparameterized for this task. However, when distilling to 17M parameters from the 170M-parameter model, intent classification error and slot error degrade by 10.95\% and 11.51\% relative to the 2.3B-parameter Stage 1 teacher.

When freezing the encoder, the Stage 2 models perform even better than the fine-tuning differential with the Stage 1 models, which is logical given the Stage 2 pretraining task. Yet again, only the 17M-parameter model cannot beat the 2.3B-parameter Stage 1 model (except for slot-filling in Domain 2).
 
Overall, Stage 2, domain-adaptive pretraining shows improved results on intent classification and slot filling tasks when compared with a model trained only on public data.

\subsection{NLU Results after Distillation} \label{sect:offlinenluresults}

We compared our distilled models to public models using the full training sets for our system (the same training data used in Section \ref{sect:onlineresults}). As public models, we consider both XLM-R Base, which has 85M non-embedding parameters, and the multilingual DistillBERT \cite{sanh2020distilbert}, which has 42M non-embedding parameters. Results are given in Table \ref{tab:offlinenlu} using exact match error rate. To count as an exact match for a given example, the model must get the intent and all slots correct.

We see that both of our distilled models outperform both public models on average. Most encouragingly, our 17M-parameter model (improvement of 4.23\% versus XLM-R) shows only minimal degradation versus our 170M-parameter model (improvement of 4.82\% versus XLM-R).

\subsection{Full System Results} \label{sect:onlineresults}

To evaluate model performance in the context of a full virtual assistant system, we follow the setup described in Section~\ref{sect:distillationsetup} and use an intermediate-sized model as a teacher-assistant to distill the final student models. 
This intermediate-sized model consisted of 170M non-embedding parameters and was distilled from a 700M-parameter Stage 1 teacher for 160K updates, the Stage 1 2.3B-parameter teacher for 105K more updates, followed by the stage 2 2.3B-parameter model for 300K more updates. See Appendix \ref{sect:hyper} for hyperparameter details of the 700M-parameter model.

\begin{table*}[h]
\centering
\begin{tabular}{lcc}
\hline
 & \textbf{Exp 1} & \textbf{Exp 2} \\
 \hline
Base Teacher Non-Embed Params & 85M & 85M \\
Base Layers/Hidden Size/FF Size & 4/312/1200 & 4/312/1200 \\
Base Non-Embed Param Count & 5M & 5M \\
Base Langs Supported & 1 & 1 \\
\hline
Cand Teacher's Teacher Non-Embed Params & 2.3B & 2.3B \\
Cand Teacher Non-Embed Params & 170M & 170M \\
Cand Layers/Hidden Size/FF Size & 4/768/1200 & 4/768/1200 \\
Cand Non-Embed Params & 17M & 17M \\
Cand Langs Supported & 9 & 9 \\
Cand Distill Examples & 80M & 200M \\
\hline
Test Locale & 1 & 2 \\
\hline
Whole System User Dissatisfaction A/B & -3.74\% & -4.91\% \\
Whole System User Dissatisfaction Tail A/B & -10.3\% & -7.50\% \\
Whole System User Dissatisfaction Seq & -14.9\% & -7.2\% \\
\hline
Offline SemER & -15.6\% & -2.98\% \\
\hline
\end{tabular}
\caption{Results from a virtual assistant experimentation platform for two experiments (Exp) in two locales comparing our candidate distilled 17M-parameter model (Cand) to baseline models (Base) distilled from an 85M-parameter teacher trained on Wikipedia data. Relative results are given for whole-system user dissatisfaction, an automatic metric, from both a parallel, A/B test with different user cohorts, as well as sequential results with the same users. Tail A/B results based on utterances not within the top 500 are also given. For reference, we also report Semantic Error Rate (SemER) for the NLU component using the same labeled test set as used for Table \ref{tab:offlinenlu}.}
\label{tab:abtest}
\end{table*}

The 170M-parameter model was then used as a teacher to distill 17M-parameter models that were used online. Before commencing distillation, the 170M-parameter teacher was fine-tuned for 15,625 updates with the same task-specific dataset that was used for the subsequent distillation process.
The distillation itself was performed using both logit matching and hidden layer matching, for which we mapped student layers (0, 1, 2, 3) to teacher layers (3, 7, 11, 15), following \citep{jiao-etal-2020-tinybert}. 
We found that optimal performance, for the two locales explored, was achieved by using two different checkpoints from the same 17M-parameter model distillation process---the first which was taken after 80M examples and the second which was taken after 200M examples.
We used a combination of 9 languages when distilling to the 17M-parameter models, being English, French, German, Hindi, Italian, Marathi, Spanish, Tamil, and Telugu.

We considered two baseline models, each being a 5M-parameter monolingual encoder distilled from a teacher with a BERT-Base architecture.
The training and distillation sets for these baseline models was comprised of Wikipedia dumps using the language in question.
The text was converted to spoken form in the same manner as described in Section \ref{sect:spokenformpartone}.
See Table \ref{tab:hyperparamsOnline} for details on architectures.

We conducted our studies using an experimentation platform for an entire virtual assistant system.
We compared our models to the baselines both in parallel, as an A/B test using a different user cohort, as well as in series using the same user cohort.
Results are given in Table \ref{tab:abtest}. 
We examined an automated measurement of user dissatisfaction across the entire virtual assistant system (not just the NLU component), which was based on the user's responses and whether the system correctly executed a task.
We also consider tail dissatisfaction, which is the dissatisfaction rate for utterances not within the top 500 most common utterances.
Finally, we provide the offline Semantic Error Rate (SemER) \cite{2020arXiv201204099P,rao2021i} results for the models using the same NLU test set as was used in Section \ref{sect:offlinenluresults}.
The SemER metric is used to evaluate the intent and slot-filling performance jointly.
Comparing a reference of tokens and their accompanying labels, performance is defined according to the following: (1) Correct slots, where the slot name and slot value is correctly identified, (2) Deletion errors, where the slot name is present in the reference but not in the hypothesis, (3) Insertion errors, where extraneous slot names are included in the hypothesis, (4) Substitution errors, where slot names from the hypothesis are included but with an incorrect slot value. Intent classification errors are substitution errors.

\begin{equation}
\textrm{SemER} = \frac{\textrm{\# Deletion + \# Insertion + \# Substitution}}{\textrm{\# Correct + \# Deletion + \# Substitution}}
\end{equation}

We find that models produced using our pretraining and distillation pipeline reduce overall user dissatisfaction by 3.74\% to 4.91\% and tail utterance dissatisfaction by 7.50\% to 10.3\% in the A/B test framework. Sequential results are even better, with up to a 14.9\% improvement, though they are less trustworthy given other possible changes to the platform over time.
Offline SemER improves by 2.98\% to 15.6\%.

One caveat of our full-system study is the difference in parameter count between the baseline models and the candidate models.
To determine the effect of final model size on performance, we fine-tuned a 5M-parameter model akin to the baselines used for experiments 1 and 2. 
We then distilled and fine-tuned an otherwise-equivalent 17M-parameter model using the same data.
Across two different languages we only saw offline SemER improvements from between 0.25\% and 0.38\% when increasing the model size from 5M to 17M parameters with everything else equal.
This suggests that a significant portion of the improvement seen in Table \ref{tab:abtest} is due to our pretraining and distillation pipeline, not due to the differing final model sizes.
Moreover, our candidate encoders were pretrained using 12 languages and distilled with 9 languages, whereas the baseline encoders were trained, distilled, and fine-tuned only with a single language.

\section{Related Work} \label{sect:related}

SLU systems, composed of both speech recognition and NLU components, have been explored extensively for decades \cite{tur2002improving,mesnil2014using,chen2019bert}.
Many recent efforts have focused on scaling up the size of pretrained language models to improve downstream tasks, including tasks related to NLU.
\cite{kaplan2020scalinglaw} proposed a power-law scaling relationship between parameter count and performance, and subsequent papers empirically confirmed this relationship for very large models, including \cite{Radford2019LanguageMA} which trained models up to 1.5 billion parameters and \cite{brown2020gpt3} which trained models up to 175 billion parameters.
Various approaches to efficiently train such large models have been explored, including model state partitioning approaches \cite{raj2019deepspeed} and pipeline parallelism approaches \cite{shoeybi2019megatronlm, harlap2018pipedream}.
Recently \cite{smith2022using} combined these approaches and trained a 530 billion parameter model.
Other lines of work have explored increasing the parameter count by introducing sparsity, using various mixture-of-expert approaches to train models of over 1 trillion parameters \cite{fedus2021switch, lewis2021base, kim2021scalable, rajbhandari2022deepspeedmoe}.

Early work \cite{10.1145/1150402.1150464, NIPS2014_ea8fcd92, 44873} suggested supervising small-sized models by using larger teacher models with the idea being that the mimicking of the teacher behavior can give small models a competitive advantage over the same-sized models trained without a teacher.
In recent years, matching internal layer outputs from student and teacher models as an auxiliary task \cite{jiao-etal-2020-tinybert, NEURIPS2020_3f5ee243, wang-etal-2021-minilmv2} has yielded even higher performance gains.

\section{Conclusion} \label{sect:conclusion}

We have described a model development pipeline in which transformer-based encoders are first pretrained from scratch using public data (Stage 1), adapted to their system using in-house, unlabeled data (Stage 2), distilled to runtime-ready sizes using a 2-step distillation process, and then fine-tuned.
Traditionally, production-focused NLU models are either distilled from models with 85M-300M parameters (Base-sized to Large-sized) and then fine-tuned, or they are trained from scratch on the final labeled dataset.
Our AlexaTM pipeline, which starts with models containing 2.3B+ parameters, significantly improves upon this paradigm, including in NLU benchmarks and in user dissatisfaction reduction across an entire virtual assistant system.
In particular, we find a large teacher, Stage 2 pretraining, a teacher-assistant distillation process, and in-domain-specific final distillation to be key techniques for improving task performance.

As future work, we would like to more robustly characterize the use of public pretrained conversational models like TOD-BERT \cite{wu2020todbert} and ConveRT \cite{henderson2020convert}, evaluate more combinations of teacher and distilled model sizes, benchmark with different public datasets like MultiATIS \cite{upadhyay2018almost,xu2020endtoend}, mTOP \cite{xia2021multilingual}, or MASSIVE \cite{fitzgerald2022massive}, make greater use of dialog and user context, experiment with code-switching, examine varying levels of ASR noise, and more.


\bibliographystyle{ACM-Reference-Format}
\balance
\bibliography{AlexaTM_KDD22}

\pagebreak

\appendix

\section{Model Hyperparameters} \label{sect:hyper}

\begin{table*}[b]
\centering
\begin{tabular}{lccc}
\hline
 Hyperparam        &    700M Teacher & 2.3B Teacher   & 9.3B Teacher   \\
\hline
 Number of Layers   & 20 & 29             & 46              \\
 Hidden size            & 1536 & 2560           & 4096            \\
 FFN inner hidden size  & 6144 & 10240          & 16384           \\
 Attention heads        & 16 & 32             & 32              \\
 Attention head size    & 64 & 80             & 128             \\
 Dropout                & 0.1 & 0.1            & 0.1             \\
 Attention Dropout      & 0.1 & 0.1            & 0.1             \\
 Warmup Steps           & 1k & 5k             & 10k             \\
 Peak Learning Rate     & 1e-3 & 1.5e-4         & 1.4e-4          \\
 Min Learning Rate      & 1e-5 & 1e-5           & 1e-5            \\
 Max Length             & 1024 & 512            & 512/1024             \\
 Batch Size (Sequences) & 2048 & 2048           & 4096            \\
 Batch Size (Tokens)    & 2M & 1M             & 2M              \\
 Weight Decay           & 0.1 & 0.1            & 0.1             \\
 LR Decay steps         & 500k & 500k           & 600k            \\
 Max Steps              &  & 950k           & 570k            \\
 Learning Rate Warmup   & Exponential & Exponential    & Linear          \\
 Learning Rate Decay    & Linear & Linear         & Linear          \\
 Adam epsilon           & 1e-8 & 1e-8           & 1e-8            \\
 Adam beta1             & 0.9 & 0.9            & 0.9             \\
 Adam beta2             & 0.99 & 0.9            & 0.99            \\
 Gradient Clipping      & 1.0 & 1.0            & 1.0             \\
\hline
\end{tabular}
\caption{Hyperparameters used for Stage 1 Teacher Models}
\label{tab:hyperparamsStage1}
\end{table*}

\begin{table*}[b]
\centering
\begin{tabular}{lcc|cccc}
\hline
 & \makecell{Base\\Teacher} & \makecell{Base\\Distillation} & \makecell{AlexaTM 170M\\Distillation Stage 1} & \makecell{AlexaTM 170M\\Distillation Stage 2} & \makecell{AlexaTM MLM\\of 170M} & \makecell{AlexaTM 17M\\Distillation} \\
 \hline
Number of Layers & 12 & 4 & 16 & 16 & 16 & 4 \\
Hidden Size & 768 & 312 & 1024 & 1024 & 1024 & 768 \\         FFN Inner Hidden Size & 3072 & 1200 & 3072 & 3072 & 3072 & 1200 \\
Attention Heads & 12 & 12 & 16 & 16 & 16 & 12 \\              Dropout & 0.1 & 0.1 & 0.1 & 0.1 & 0.1 & 0.1 \\                Peak Learning Rate & 0.5 & 0.5 & 1.50E-03 & 2.00E-04 & 1.00E-05 & 1.00E-03 \\
LR Warmup Type & Noam & Noam & Exponential & Exponential & Exponential & Exponential \\
LR Decay & Noam & Noam & Linear & Linear & Linear & Linear \\
Warmup Steps & 3250 & 3250 & 100k & 10k & 10k & 500 \\        LR Decay Steps & N/A & N/A & 1M & 250k & 15.6k & 195k \\
Max Length & 512 & 512 & 512 & 30 & 512 & 512 \\
Tokens per Batch & 64k & 128k & 393k & 492k & 24.6k & 8192 \\
Number of Updates & 40k & 40k & 1.5M & 360k & 1M & 5M / 12.5M \\                                                            Adam Epsilon & 1.00E-09 & 1.00E-09 &  &  &  &  \\
Adam Beta1 & 0.9 & 0.9 &  &  &  &  \\                         Adam Beta2 & 0.98 & 0.98 &  &  &  &  \\
Lamb Epsilon &  &  & 1.00E-08 & 1.00E-08 & 1.00E-08 & 1.00E-08 \\
Lamb Beta1 &  &  & 0.9 & 0.9 & 0.9 & 0.9 \\
Lamb Beta2 &  &  & 0.999 & 0.999 & 0.999 & 0.999 \\
\hline
\end{tabular}
\caption{Hyperparameters used for full-system experiments described in Section \ref{sect:onlineresults}}.
\label{tab:hyperparamsOnline}
\end{table*}

See Table \ref{tab:hyperparamsStage1} for Stage 1 teacher model hyperparameters and Table \ref{tab:hyperparamsOnline} for hyperparemeters used with models associated with our full-system experiments (Section \ref{sect:onlineresults}).

\end{document}